\documentclass{optica-article}

\journal{opticajournal} % for journals or Optica Open

\articletype{Research Article}

\usepackage{booktabs}
\usepackage{multirow}
\begin{document}

\title{Video Frame Interpolation for Polarization via Swin-Transformer}

\author{Feng Huang,\authormark{1} Xin Zhang,\authormark{1} Yixuan Xu,\authormark{1} Xuesong Wang\authormark{1} and Xianyu Wu\authormark{1*}}

\address{\authormark{1} with the School of Mechanical Engineering and Automation, Fuzhou University, Fujian, 35018, China. }

\email{\authormark{*}xwu@fzu.edu.cn} %% email address is required; see note below about the corresponding author designation

% use {asbstract*} to suppress the copyright line. Copyright information will be added in production

\begin{abstract*} 
Video Frame Interpolation (VFI) has been extensively explored and demonstrated, yet its application to polarization remains largely unexplored. Due to the selective transmission of light by polarized filters, longer exposure times are typically required to ensure sufficient light intensity, which consequently lower the temporal sample rates. Furthermore, because polarization reflected by objects varies with shooting perspective, focusing solely on estimating pixel displacement is insufficient to accurately reconstruct the intermediate polarization. To tackle these challenges, this study proposes a multi-stage and multi-scale network called Swin-VFI based on the Swin-Transformer and introduces a tailored loss function to facilitate the network's understanding of polarization changes. To ensure the practicality of our proposed method, this study evaluates its interpolated frames in Shape from Polarization (SfP) and Human Shape Reconstruction tasks, comparing them with other state-of-the-art methods such as CAIN, FLAVR, and VFIT. Experimental results demonstrate our approach's superior reconstruction accuracy across all tasks.
\end{abstract*}
\section{Introduction}
\label{Introduction}
\begin{figure}[!h]
\centering
\includegraphics[width=3.5in]{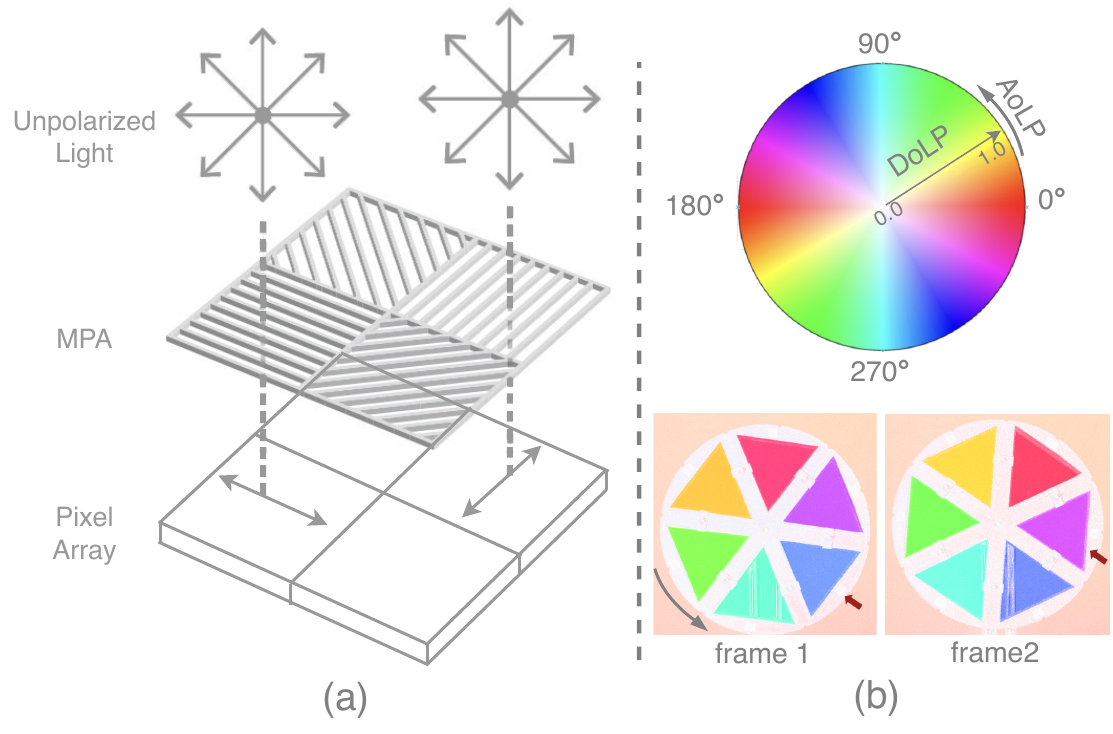}
\caption{The necessity and challenge of VFI for polarization. (a) The intensity of polarized light is much weaker after passing through micro-polarizer array of a DoFP polarimeter. (b) Upper: AoLP-DoLP visualization, where AoLP and DoLP are mapped to hue and brightness, respectively. Lower: Variation of AoLP induced by alterations of the shooting perspective, posing a challenge for polarized video frame interpolation. (Note the polarizer's AoLP change indicated by the arrow during the rotation process)}
\label{chanllenge}
\end{figure}

Polarization, as a fundamental property of light extending beyond intensity and wavelength, carries crucial information about objects, including surface roughness, three-dimensional (3D) normals, and material composition. Its significance is underscored by its widespread applications in diverse fields such as 3D imaging\cite{ba2020deep,zou2022human,zou2020detailed}, object recognition\cite{usmani2021deep}, and biomedical imaging\cite{li2021polaromics}. In recent years,  polarization imagers based on division-of-focal-plane (DoFP) have received significant attention due to their ability to capture different polarization states in real time. As depicted in Fig. \ref{chanllenge}(a). a Micro-Polarizer Array (MPA) is integrated with a DoFP polarization imaging sensor to allow light waves oscillating in the particular orientations such as $0^{\circ}, 45^{\circ}, 90^{\circ}, 135^{\circ}$ to pass through, which makes the intensity of polarized light much weaker. To ensure the pixel array receive sufficient polarized light, slower shutter speeds are typically required, which consequently lead to issues such as reduced temporal sample rates and motion blur. The application of polarized light heavily relies on the quality of polarization imaging, which underscores the necessity for VFI in the context of polarization.

In order to further analyze the potential issues inherent in VFI for Polarization, this work collects a dataset, named PVFI-mono, featuring simple scenes (translation and rotation) with strong polarization characteristics to mitigate potential confounding factors. Our research finds that frame interpolation for polarization presents is rather challenging. As illustrated in the bottom of Fig. \ref{chanllenge}(b), the polarization information of an object's reflection changes with variations in the camera's shooting perspective. Six polarizers are mounted on a circular bracket, with each adjacent polarizer differing by 30 degrees. These polarizer regions exhibit a strong Degree of Linear Polarization (DoLP)\cite{zhao2016multi}. In the two consecutive frames, the red arrow points to the same polarizer. It can be observed that during rotation, the Angle of Linear polarization (AoLP)\cite{zhao2016multi} of the same polarizer changes (refer to top of Fig. \ref{chanllenge}(b), where AoLP and DoLP are mapped to hue and brightness, respectively). This indicates that as each polarizer rotates, not only does the pixel position shift, but the polarization information also changes. In essence, unlike conventional VFI, which solely addresses pixel displacements, VFI for polarization requires a comprehensive understanding of the variation of polarization information, such as AoLP, induced by fluctuations in camera shooting perspective.

Video frame interpolation is the process of reconstructing uncaptured intermediate frames during the exposure time by synthesizing adjacent frames, which can enhance its visual quality and smoothness of motion. Existing methods can be mainly classified into flow-based methods and kernel-based methods. 
Calculating the optical flow of polarized light is a challenging task, and thus, flow-based methods are not considered within the scope of this work. Kernel based methods\cite{niklaus2017video,cheng2020video,lee2020adacof,choi2020channel,choi2021multi,kalluri2023flavr,shi2022video,cheng2019multi} have gained increasing popularity in recent years, which primarily focus on expanding the receptive field of convolutional kernels. Despite significant achievements, these methods still lack the ability to effectively establish long-range correlations.
Meanwhile, Transformers \cite{vaswani2017attention} have demonstrated its great potential in various image tasks such as image classification \cite{dosovitskiy2020image,xie2018rethinking}, object detection \cite{misra2021end,carion2020end}, and image restoration \cite{liang2021swinir,zamir2022restormer}, due to their ability to capture long-range dependencies and contextual relationships in sequences. However, extending the Transformer to video tasks is not as straightforward as extending 2D convolutions to 3D convolutions, as it poses challenges such as computational complexity and memory requirements.

\begin{figure*}[!h]
\centering
\includegraphics[width=1\textwidth]{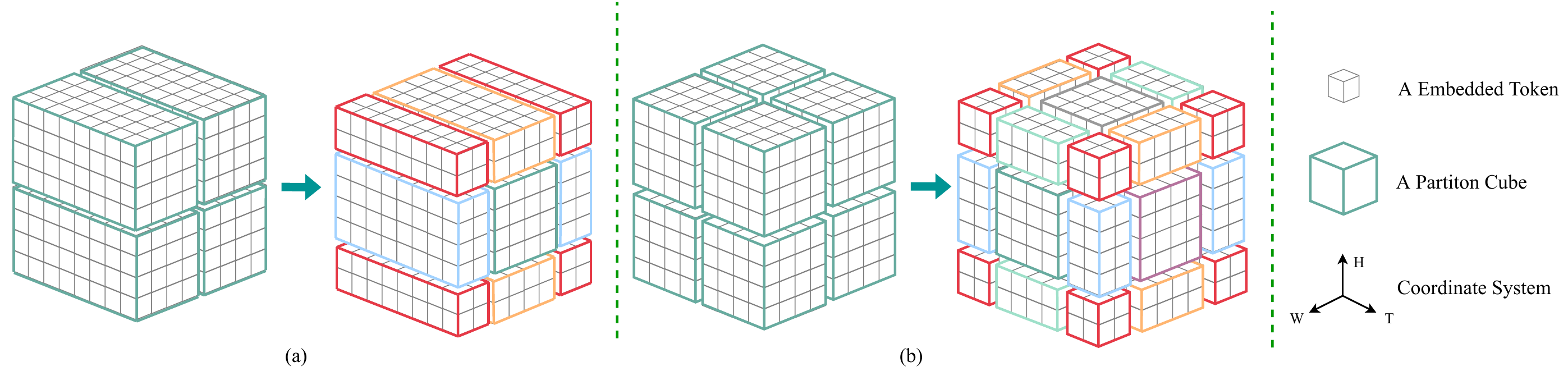}
\caption{ (a) A naive expansion of Swin-Transformer to spatial-temporal space. (b) An simple illustration of Swin-VFI, where the boundaries of the dimensions are connected, and the cubes with the same color are merged and masked after being shifted.}
\label{Swin-VFI}
\end{figure*}

In this work, Video Swin-Transformer\cite{liu2022video} is introduced into VFI for polarization task, namely Swin-VFI. Compared with the naive expansion of Swin\cite{liu2021swin} to spatial-temporal space shown in Fig. \ref{Swin-VFI}(a), Swin-VFI shown in Fig. \ref{Swin-VFI}(b) treats 3D-patches as tokens and partitions them into cubes with a fixed size along the height, width, and time axis. Local self-attention is computed within each cube, followed by shifted-cube mechanism to establish connections between adjacent cubes. This approach enables the model to exploit spatiotemporal locality inductive bias and exhibit linear computational complexity with respect to the patch number, making it concise, efficient, and demonstrating exceptional performance. 

In summary, the main contributions of our work are as follows:
\begin{itemize}

\item For the task of VFI for polarization, this work collects a dataset with strong polarization characteristics to analyze its necessity and challenge. Based on a local self-attention mechanism, a multi-stage multi-scale Transformer named Swin-VFI is introduced to establish long-range correlations within videos. Additionally, this work propose a tailored loss to enhance the accuracy of reconstructing polarization information. To the best of our knowledge, this is the first task based on a data-driven approach for VFI in the context of polarization.

\item This work trains CAIN\cite{choi2020channel}, FLAVR\cite{kalluri2023flavr}, VFIT-S\cite{lu2022video}, VFIT-B\cite{lu2022video}, and our proposed Swin-VFI on the PVFI-mono dataset and the PHSPD dataset. Quantitative and qualitative experiments demonstrate that our Swin-VFI dramatically outperforms state-of-the-art methods, showing more accurate reconstruction accuracy of intensity and polarization information while requiring much cheaper Params and FLOPS.

\item This study conducts tests on the interpolated frames generated by our proposed method and other state-of-the-art approaches in the SfP task and Human Shape Reconstruction task. The experimental results demonstrate that our method exhibits superior estimation in surface normals and human shape and pose.  

\item Our Swin-VFI also performs well on conventional VFI task, achieving the best results on Vimeo-90K, DAVIS, SNU-FILM and Xiph datasets, with a reduction of about 40\% in the number of parameters compared to VFIT-B.
\end{itemize}

\section{Related Works}

\subsection{Video Frame Interpolation}

The objective of VFI is to generate intermediate frames by combining adjacent frames that were not captured during the exposure period. This longstanding and classical problem in video processing is currently tackled through three prominent approaches: phase-based methods, optical flow-based methods, and kernel-based methods.

\subsubsection{Phase-based methods} Phase-based methods\cite{meyer2015phase,meyer2018phasenet} utilize Fourier theory to estimate motion by analyzing the phase discrepancy between corresponding pixels in consecutive frames. These techniques generate intermediate frames by applying phase-shifted sinusoids. However, the 2$\pi$-ambiguity problem can pose a significant challenge in determining the correct motion.

\subsubsection{Flow-based methods} Flow-based methods\cite{bao2019depth,lee2020adacof,liu2017video,xu2019quadratic,jiang2018super,park2020bmbc,lu2022video,choi2018triple,rufenacht2018temporal} utilize optical flow estimation to perceive motion information and capture dense pixel correspondence between frames. These methods use a flow prediction network to compute bidirectional optical flow that guides frame synthesis, along with predicting occlusion masks or depth maps to reason about occlusions. 
However, flow-based methods are not within our scope of consideration because estimating optical flow for polarized light is a rather challenging task, as it necessitates the consideration of both intensity and polarization data.

\subsubsection{Kernel-based methods} Kernel-based methods have gained momentum in VFI since the emergence of AdaConv \cite{niklaus2017vvideo}, a method that uses a fully convolutional network to estimate spatially adaptive convolution kernels. This is because it no longer requires motion estimation or pixel synthesis like flow-based methods. DSepConv \cite{cheng2020video} and AdaCoF \cite{lee2020adacof} employ Deformable convolution to overcome the limitation of a fixed grid of locations in original convolution. CAIN \cite{choi2020channel} expands the receptive field size of convolution by utilizing Pixel Shuffle. SepConv \cite{niklaus2017video} performs separable convolution, thereby reducing the Params and memory usage. Then, FLAVR \cite{kalluri2023flavr} substitutes the 2D convolutions utilized in Unet with their 3D counterparts, while applying feature gating to each of the resultant 3D feature maps. This achieves the best performance among CNN-based methods at the cost of a large Params. However, these CNN-based architectures still cannot overcome their inherent limitation of using fixed-size kernels, which prevent them from capturing global dependencies to handle large motion and limit their further development for VFI task. Inspired by Depth-wise separable convolution \cite{howard2017mobilenets}, Zhihao Shi et al. introduce VFIT \cite{shi2022video}, a separated spatio-temporal multi-head self-attention mechanism, which outperforms all existing CNN-based approaches while significantly reducing the Params. Within the field of kernel-based methods, CNN backbones have undergone a developmental trajectory from 2D to separable and then to 3D kernels. Zhihao Shi et al. has proposed a space-time separation strategy \cite{shi2022video} in Transformer methods. In this work, a 3D version of the local self-attention mechanism  is introduced to the VFI task.
\subsection{Vision Transformer}
The key innovation of the ViT \cite{dosovitskiy2020image} is its application of the Transformer architecture, originally developed for natural language processing, to computer vision tasks. This represents a notable departure from the standard backbone architecture of CNNs in computer vision. By dividing the image into a sequence of patches and leveraging the Transformer encoder to capture global dependencies between them, ViT achieves impressive performance on image classification benchmarks. This pioneering work has paved the way for subsequent research aimed at improving the utility of the ViT model, and underscores the potential of the Transformer architecture in computer vision applications. To mitigate the computational and memory challenges associated with ViT, Swin-Transformer \cite{liu2021swin} partitions the embedded patches into non-overlapping windows. Within each window, local self-attention is calculated by ViT. Subsequently, shifted-window self-attention is computed to establish the correlation among windows. This strategy has demonstrated remarkable performance in various image tasks, such as image classification \cite{liu2021swin,bertasius2021space}, object detection \cite{carion2020end,misra2021end}, and image restoration \cite{chen2021pre,zamir2022restormer}, achieving state-of-the-art results. Despite Swin-Transformer's success in image tasks, extending it to video tasks by simply expanding along the time dimension resurfaces thorny computational and memory issues \cite{shi2022video}. To address these issues, Ze Liu et al. further proposed a new 3D shifted windows mechanism \cite{liu2022video} that efficiently captures temporal information, reduces the computational and memory demands, and achieves state-of-the-art results on video action recognition tasks. This method makes Swin-Transformer a promising approach for video analysis tasks. 

\subsection{Shape from Polarization}
Shape from Polarization is a technique utilized for reconstructing the 3D shape or surface geometry of an object based on the polarization properties of light. Unlike traditional methods relying on intensity and color information, SfP leverages changes in polarized light as it interacts with surfaces from various angles and orientations. Previous approaches often incorporate physical priors like coarse depth maps\cite{yang2018polarimetric} and smooth object surfaces\cite{atkinson2006recovery} to address inherent ambiguities. A primary challenge in SfP is the $\pi$-ambiguity between the azimuth angle and polarization intensity. To mitigate this issue, \cite{ba2020deep} introduced a data-driven method, employing physical priors as inputs to train a deep learning model, significantly reducing shape error. Furthermore, \cite{zou2022human,zou2020detailed} collected a Polarization Human Shape and Pose Dataset (PHSPD), and classified ambiguous normals and background for each pixel and regressed the normals based on the ambiguous and classified physical priors, further refining the technique.

\section{PROPOSED METHOD}
\subsection{Polarization Imaging Mechanism}

In section \ref{Introduction}, figure \ref{chanllenge}(a) depicted the mechanism of DoFP polarization imaging sensor. Each pixel of the sensor captures polarized light at one of four orientations: $0^{\circ}, 45^{\circ}, 90^{\circ}, 135^{\circ}$, enabling four channels real-time synchronized polarization imaging and precise polarization analysis in various applications. The four intensity measurements $I(\theta)$ of the light wave filtered by a linear polarizer oriented at $\theta (\theta = 0^{\circ}, 45^{\circ}, 90^{\circ}, 135^{\circ})$ from the DoFP sensor super-pixel are used to calculate the Stokes parameters, which can be expressed as follows \cite{thompson2003polarized}: 

\begin{equation}
\begin{aligned}
& \mathcal{S}_0=0.5\left(I\left(0^{\circ}\right)+I\left(45^{\circ}\right)+I\left(90^{\circ}\right)+I\left(135^{\circ}\right)\right) \\
& \mathcal{S}_1=I\left(0^{\circ}\right)-I\left(90^{\circ}\right) \\
& \mathcal{S}_2=I\left(45^{\circ}\right)-I\left(135^{\circ}\right)
\end{aligned}
\label{stokes}
\end{equation}
To observe polarization, DoLP and AoLP are of most significance, defined as \cite{zhao2016multi} :
\begin{equation}
\begin{aligned}
& D o L P=\frac{\sqrt{\mathcal{S}_1^2+\mathcal{S}_2^2}}{S_0} \\
& A o L P=\frac{1}{2} \tan ^{-1}\left(\frac{\mathcal{S}_2}{\mathcal{S}_1}\right)
\end{aligned}
\label{adolp}
\end{equation}

\subsection{Structure of the neural network}
\begin{figure*}[!h]
\centering
\includegraphics[width=1\textwidth]{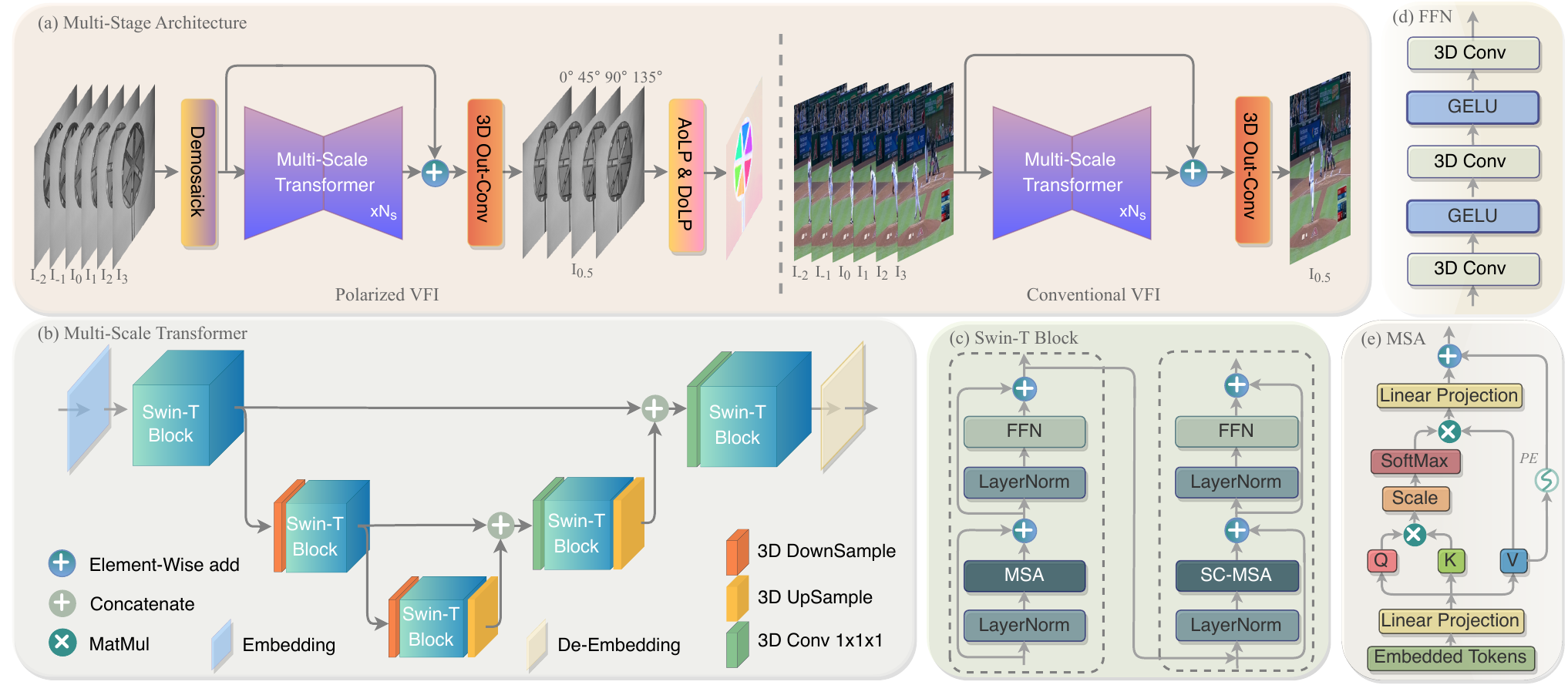}
\caption{The overall pipeline of Swin-VFI. (a) Multi-stage Architecture. (b) Multi-scale Transformer. (c) An illustration of Swin Transformer blocks, which contains two successive Multi-head Self-Attention blocks. (d) Feed Forward Network. (e) Brief explanation of Multi-head Self-Attention.}
\label{arch}
\end{figure*}
Fig. ~\ref{arch}(a) depicts our multi-stage architecture that utilizes $N_s$ cascaded Multi-Scale Transformer for polarized VFI and conventional VFI. Instead of selecting two or four adjacent frames next to the reference frame as input as \cite{choi2020channel,kalluri2023flavr,shi2022video}, our approach chooses six adjacent frames to accurately estimate the motion of the interpolated frame.  Moreover, a long identity mapping is employed to mitigate the vanishing gradient problem. The desired interpolated frame is finally obtained via a 3D convolution operation.
Fig. ~\ref{arch}(b) illustrates the network structure of the Multi-Scale Transformer when $N_s=1$. Specifically, the embedding layer converts the input frames into dense representations. The de-embedding layer performs the inverse operation. In order to enable multi-scale self-attention, it is crucial to downsample the output of the Swin-T block from the preceding scale before each Swin-T block in the encoder. Likewise, the Decoder performs the inverse operation. Moreover, skip connections are employed at same scale, while a $1\times1\times1$ convolutional operation is applied to halve the depth of the concatenated feature maps. The concise procedure of Swin-T is portrayed in Fig. ~\ref{arch}(c), where as the Multi-Head Self Attention (MSA) module and feed-forward network (FFN) are illustrated in Fig. ~\ref{arch}(d). and Fig. ~\ref{arch}(e) respectively. 

\subsection{Swin-VFI}

According to the findings presented in  \cite{shi2022video}, the straightforward 3D extension of the Swin Transformer, i.e., the Spatial-Temporal Swin (STS) attention layer, illustrated in Fig. \ref{Swin-VFI} (a), can efficiently lower the computational complexity of the Global MSA by evenly dividing the input tensor $X_{in} \in \mathbb{R}^{T \times H\times W \times D}$ into $\frac{HW}{hw}$ non-overlapping cubes. However, this  approach has not yet achieved an easily trainable level. In light of this, this work introduces the application of the Video Swin Transformer \cite{liu2022video}  to the VFI task, as depicted in Fig. \ref{Swin-VFI} (b).
In accordance with \cite{liu2022video}, the input tensor $X_{in}\in \mathbb{R}^{T\times H\times W \times D}$ is subjected to a process of partitioning into $\frac{THW}{thw}$ non-overlapping cubes of size $t\times h\times w$ utilizing an even partitioning strategy, as presented in Fig. \ref{Swin-VFI} (b).  The resulting cubes are reshaped into $X\in \mathbb{R}^{THW\times D }$ by the Swin-VFI module. Linear transformations, specifically $W_q$, $W_k$, and $W_v\in \mathbb{R}^{D\times D}$, are employed to produce the query $Q$, key $K$, and value $V\in \mathbb{R}^{THW\times D}$ representations, respectively.

\begin{equation}
Q=XW_q,K=XW_k,V=XW_v
\label{qkv}
\end{equation}

Similarly, $Q,K,V\in \mathbb{R}^{THW \times D}$ are divided into $n$ heads: $Q = [Q_1,...,Q_n]$, $K = [K_1,...,K_n]$, $V = [V_1,...,V_n]$, so that each head has a dimension of $d_h = {\frac{D}{n}}$. The multi-head self-attention operation is then conducted within each cube according to the following equation:
\begin{equation}
\begin{cases}
Swin-VFI(Q_j,K_j,V_j,d) = [\mathop{Concat}\limits_{j=1}^n(head_j )]W + P(v)\\head_j =softmax({\frac{Q_jk^T_j}{d}})V_j    
\end{cases}
\label{Swin-VFI_equation}
\end{equation}

where $d \in \mathbb{R}^{1}$ and $W \in \mathbb{R}^{D \times D}$ are learnable parameters. $P(V)=3DConv(Gelu(3DCon(V)))$  is the function to generate positional embedding. To establish connections among the cubes,  as depicted in Fig. \ref{Swin-VFI} (b), each cube is shifted along the time, height, and width dimensions by $t/2$, $h/2$ and $w/2$ steps, respectively. The Shifted Cube Multi-head Self-Attention (SC-MSA) is calculated within each new cube.

The process in Eq. (\ref{qkv}) and Eq. (\ref{Swin-VFI_equation}) is calculated for $\frac{THW}{thw}$ times, and its computational complexity can be specifically expressed as:
 
\begin{equation}
\Omega(Swin-VFI) = {4D^2(THW)}+2thwD(THW)
\label{Swin-VFI-MSA}
\end{equation}

 Thanks to the local self-attention mechanism, Swin-VFI reduces the computational complexity from a quadratic relationship with the number of patches for ViT \cite{dosovitskiy2020image} to a linear relationship. 

\subsection{Video Frame Interpolation for Polarized Video}
\label{PVFI}

\subsubsection{PVFI-Mono Dataset}

\begin{figure}[ht]
    \centerline{\includegraphics[width=3.5in]{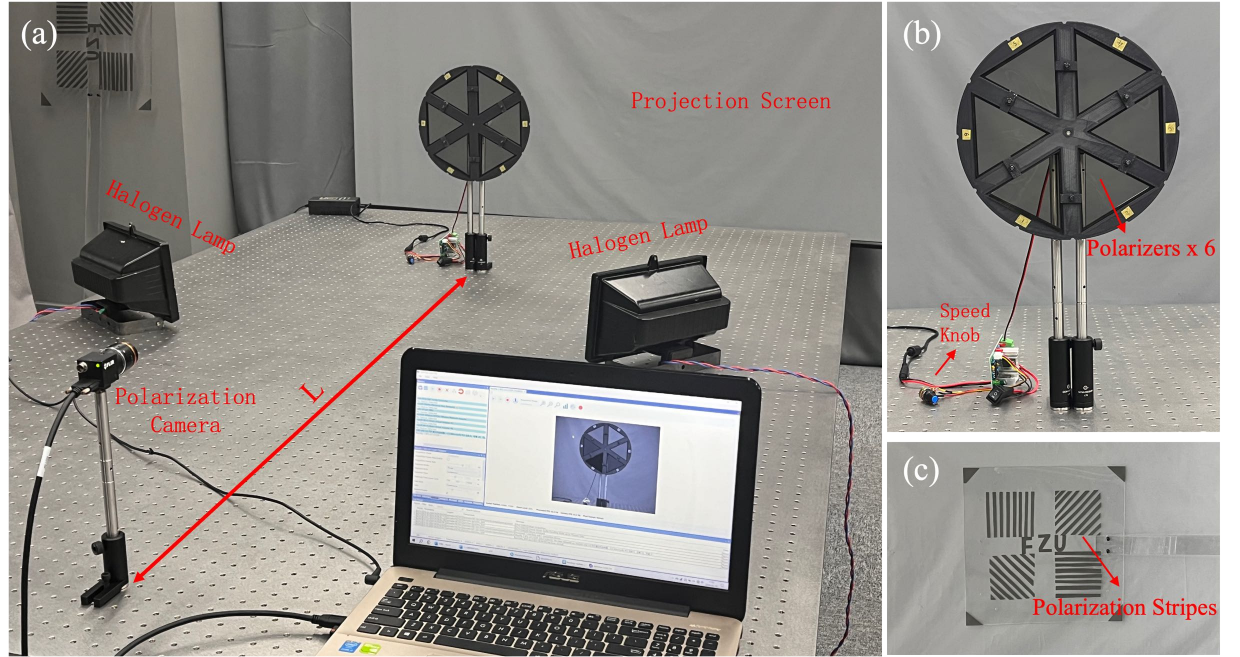}}
     \caption{(a) The capture scene of the PVFI-Mono dataset. (b) The shooting target in the rotation scenario. (c) The shooting target in the translation scenario.}
    \label{collect}
\end{figure}

This study collect a dataset for polarized video frame interpolation, named PVFI-Mono. The capture scene is illustrated in Fig. \ref{collect}(a). The camera used is FLIR BFS-U3-51S5P with a focal length of 25mm, and resolution of $2048\times2448$ (with momentary field of view loss). To eliminate unnecessary interference, a projection screen that reflects unpolarized light is used as the background. The shooting scenes include rotation and translation. The shooting target in the rotation scenario is shown in Fig. \ref{collect}(b), where polarizers are installed on a circular frame, with the angles of polarizers increasing sequentially by $30^{\circ}$ in order. The circular frame is mounted on a variable-speed DC motor (maximum idle speed of 300 rpm), and halogen lamps are used to increase scene brightness and mitigate severe dynamic blur caused by high motor speed. Twenty sets of data are collected by adjusting the motor speed and the distance $L$ between the focal plane of the camera and the plane where the polarizer is located. The shooting object in the translation scenario is shown in Fig. \ref{collect}(c), where several polarizers are pasted on a transparent acrylic board to create stripes. The angles of the polarizers match the direction of the stripes. The shooting object is moved by hand-held movement, with random adjustments in distance and speed during movement. Twenty sets of data are similarly collected. 

\subsubsection{Loss Function}

To improve the accuracy of polarization information reconstruction, considering the variations in AoLP caused by changes in camera viewing angles, in addition to the intensity loss described in Eq. \ref{L_I}, this study introduces the Stokes loss outlined in Eq. \ref{L_S}, the AoLP loss indicated in Eq. \ref{L_A}, and the DoLP loss specified in Eq. \ref{L_D} for the VFI for polarization task.
\begin{equation}
\mathcal{L}_{\mathrm{I}}=\left\|{\mathcal{F}(\mathbf{I}}_{-2},\mathbf{I}_{-1},\mathbf{I}_{0},\mathbf{I}_{1},\mathbf{I}_{2},\mathbf{I}_{3})-\mathbf{I}_{0.5}\right\|_1 
\label{L_I}
\end{equation}
\begin{equation}
\mathcal{L}_{\mathcal{S}}=\frac{1}{3} \sum_{i=0}^{2}\left\|\mathcal{S}_i\left(\mathcal{F}(\mathbf{I}_{-2},\mathbf{I}_{-1},\mathbf{I}_{0},\mathbf{I}_{1},\mathbf{I}_{2},\mathbf{I}_{3}\right)-\mathcal{S}_i(\mathbf{I}_{0.5})\right\|_2
\label{L_S}
\end{equation}
\begin{equation}
\mathcal{L}_{\mathcal{A}}=\left\|\mathcal{A}_i\left(\mathcal{F}(\mathbf{I}_{-2},\mathbf{I}_{-1},\mathbf{I}_{0},\mathbf{I}_{1},\mathbf{I}_{2},\mathbf{I}_{3}\right)-\mathcal{A}_i(\mathbf{I}_{0.5})\right\|_2
\label{L_A}
\end{equation}
\begin{equation}
\mathcal{L}_{\mathcal{D}}=\left\|\mathcal{D}_i\left(\mathcal{F}(\mathbf{I}_{-2},\mathbf{I}_{-1},\mathbf{I}_{0},\mathbf{I}_{1},\mathbf{I}_{2},\mathbf{I}_{3}\right)-\mathcal{D}_i(\mathbf{I}_{0.5})\right\|_2
\label{L_D}
\end{equation}
where $\mathcal{F}(\cdot)$ represents our neural network, $\mathcal{S}_i(\cdot)$ denotes the computation of Stokes parameters \cite{tyo2006review} as formulated in Eq.\ref{stokes}, $\mathcal{A}_i(\cdot)$ and $\mathcal{D}_i(\cdot)$ represent the computation of AoLP and DoLP expressed in Eq. \ref{adolp}, the process of demosaicking\cite{polanalyser}
 and normalization is omitted for the sake of brevity. The final loss for polarized data is obtained as follows:
\begin{equation}
\mathcal{L} = \lambda_1\mathcal{L}_I + \lambda_2\mathcal{L}_S + \lambda_3\mathcal{L}_A + \lambda_4\mathcal{L}_D
\label{L}
\end{equation}
where $\lambda_1$, $\lambda_2$, $\lambda_3$, and $\lambda_4$ are the weights corresponding to the different loss terms. Their configuration adheres to the principle that the differences between intensity loss and polarization loss should be within an order of magnitude (closer values are preferred). Building upon the constraint of intensity loss $\mathcal{L}_I$, our research further selects one constraint from $\mathcal{L}_S$, $\mathcal{L}_A$, and $\mathcal{L}_D$ to form the loss function for training Swin-VFI on the PVFI-Mono dataset. The results of this training process are detailed in Section \ref{loss}. Ultimately, the values are set as $\lambda_1=0.1$, $\lambda_2=1$, and $\lambda_3=\lambda_4=0$.

\section{EXPERIMENTAL RESULTS}
\subsection{Implementation Details}
\subsubsection{Training}Consistent with \cite{shi2022video}, the training batch size is set to 4. The Adam optimizer \cite{kingma2014adam} is utilized with $\beta_1= 0.9$ and $\beta_2= 0.99$. The learning rate is initialized to $2e^{-4}$, and a Cosine Annealing scheme is adopted over 100 epochs. For polarized data, the loss function $\mathcal{L}$ defined in Eq. \ref{L} is utilized, whereas for non-polarized data, the intensity loss $\mathcal{L}_{\mathrm{I}}$ outlined in Eq. \ref{L_I} is employed. The input frames are randomly corped with size of $128 \times 128$; The stage number $N_s$ for the polarized model is configured as 1 for the purpose of optimizing computational efficiency, while in the case of the non-polarized model, it is set to 3. For data augmentation, this work follows the method used in FLAVR \cite{kalluri2023flavr}, which randomly apply horizontal and vertical flips and temporal flips to the input sequence.

\subsubsection{Dataset} On the one hand, for polarized data, our model is trained on the PVFI-Mono dataset which contains 1014 septuplets, with 816 for training and 198 for testing. Moreover, our research further trained our model on PHSPD dataset\cite{zou2020polarization}, a collection designed for 3D human shape reconstruction from polarization images. This dataset synchronizes four cameras, comprising one polarization camera and three Kinects v2, capturing views from different angles. To ensure the imaging quality of the polarization camera, the frame rate of this synchronization system is limited to 15 frames per second (fps). This study extracts the polarization data from this dataset for use in polarized video frame interpolation. The dataset consists of 4057 septuplets, with 3245 samples allocated for training and 812 for testing. On the other hand, our non-polarized model is trained on Vimeo-90K septuplet training set \cite{xue2019video}, which includes 64,612 seven-frame sequences with a resolution of $448\times 256$.  

The performance of our models is accessed on widely-used datasets, including the Vimeo-90K septuplet test set \cite{xue2019video}, which comprises 7824 septuplets with a resolution of $448 \times 256$; the DAVIS dataset \cite{perazzi2016benchmark}, containing 2849 triplets with a resolution of $832 \times 448$; SNU-FILM dataset \cite{choi2020channel}, which is classified four categories based on the degree of motion: Easy, Medium, Hard, and Extreme. Each category comprises 310 triplets, primarily with a resolution of $1280 \times 720$;  and the Xiph dataset\cite{kong2022ifrnet}, which is downsampled and center-corped to get “Xiph-2K” and “Xiph-4K” subset. This study transform the DAVIS\cite{perazzi2016benchmark}, SNU-FILM\cite{choi2020channel}, and Xiph\cite{kong2022ifrnet} datasets into septuplet testsets to conduct comprehensive comparisons.

\subsection{Evaluation}

\subsubsection{Comparisons on the polarized datasets}

\begin{figure*}[!h]
\centering
\includegraphics[width=1\textwidth]{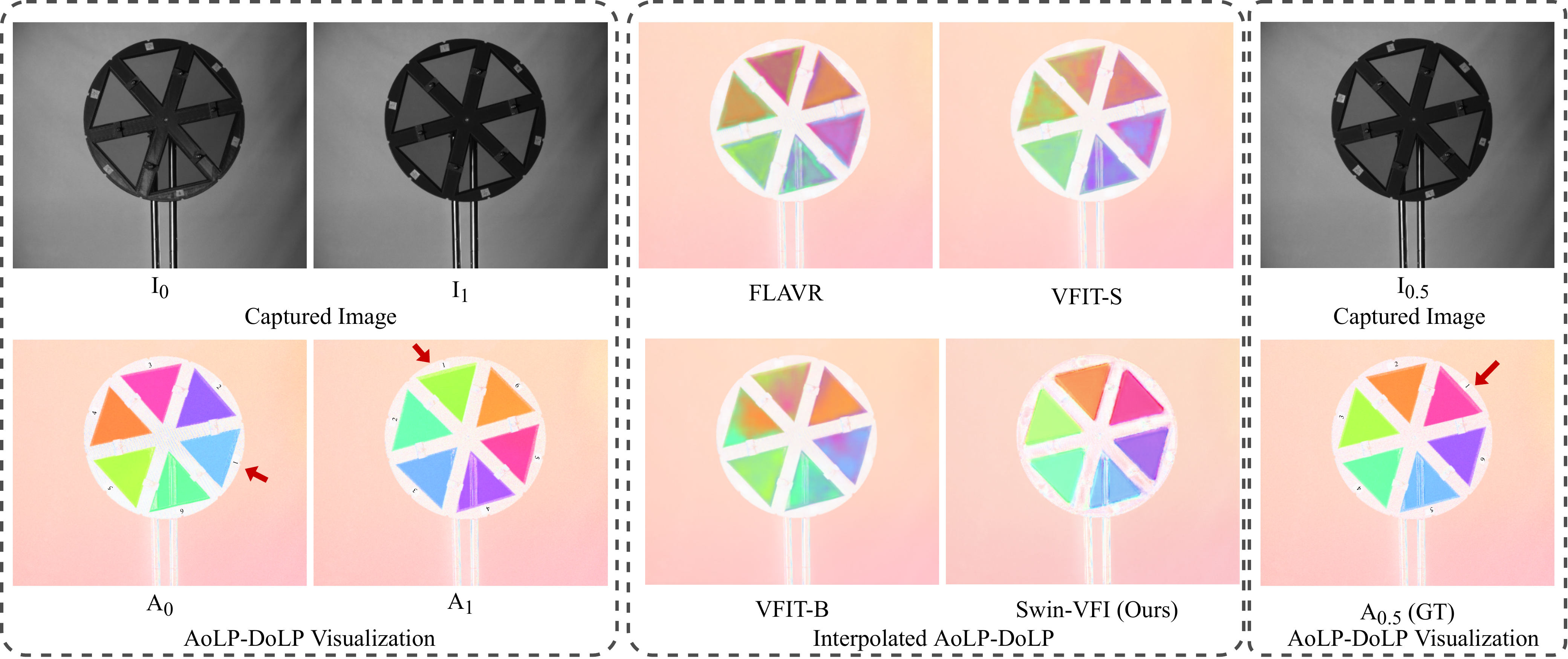}
\caption{(a): Captured images $\mathrm{I}_{\text{0}}, \mathrm{I}_{\text{1}}$ and their corresponding AoLP and DoLP visualizations $\mathrm{A_0, A_1}$. (The red arrow points to the same polarizer.) (b): Visualizations of AoLP and DoLP obtained from the pretrained models provided by FLAVR\cite{kalluri2023flavr}, VFIT-S \cite{shi2022video}, VFIT-B \cite{shi2022video} , as well as our proposed Swin-VFI method. (c): Visualizations of AoLP and DoLP $\mathrm{A_{0.5}}$ for the intermediate frame $\mathrm{I_{0.5}}$ between $\mathrm{I_0}$ and $\mathrm{I_1}$.}
 
\label{rotating}
\end{figure*}

Our research first attempts to compute the polarization information of the intermediate frame using the pretrained models provided by traditional VFI methods such as FLAVR, VFIT-S, and VFIT-B, and compared these results with our proposed Swin-VFI method. As shown in the Fig.\ref{rotating}, traditional VFI methods fail to accurately reconstruct AoLP. This is because traditional methods focus solely on the displacement of pixels between frames, whereas polarization information changes with variations in the camera's viewing angle. Consequently, traditional VFI methods cannot be directly applied to polarized data. Our proposed Swin-VFI method, specifically trained on polarized data and incorporating polarization loss terms, reconstructs results that are highly similar to the AoLP and DoLP visualization of the intermediate frame.

\begin{table*}
    \caption{Quantitative comparisons on the PVFI-Mono dataset.}
    \label{PVFI-Mono comparisons}
    \centering
    \begin{tabular*}{1\textwidth}{@{\extracolsep{\fill}}@{\hspace{5pt}}c@{\hspace{5pt}}c@{\hspace{5pt}}c@{\hspace{5pt}}c@{\hspace{5pt}}c@{\hspace{5pt}}c@{\hspace{5pt}}c@{\hspace{5pt}}c@{\hspace{5pt}}c@{\hspace{5pt}}c@{\hspace{5pt}}c@{\hspace{5pt}}c@{\hspace{5pt}}c@{\hspace{5pt}}c@{\hspace{5pt}}}
        \toprule
        \multirow{2}{*}{Methods} &\multirow{2}{*}{Params (M)} &\multirow{2}{*}{FLOPS (G)} &\multicolumn{4}{c}{PVFI-Mono}\\
        \cmidrule(r){4-7} \cmidrule(r){8-11} 
         &&&Intensity		&Stokes		&AoLP		&DoLP \\
        \midrule
        CAIN \cite{choi2020channel}    &78.40 &\underline{19.94} &26.96/0.849 &30.59/0.941 &9.90/0.281 &21.02/0.682\\
        FLAVR \cite{kalluri2023flavr}  &42.06 &99.98 &\underline{27.85/0.877} &\underline{31.45/0.952} &\textbf{11.47/0.308} &23.03/0.730 \\
        VFIT-S \cite{shi2022video}     &\underline{7.54}     &29.83   &27.25/0.885 &31.11/0.953 &11.00/0.313 &\textbf{23.68/0.759}  \\
        VFIT-B \cite{shi2022video}     &29.08 &63.59&27.41/0.884 &31.24/0.953 &\underline{11.02/0.313} &\underline{23.56/0.757}      \\
        Swin-VFI                       &\textbf{4.13}   &\textbf{18.33} &\textbf{28.17/0.880} &\textbf{31.91/0.953} &10.28/0.297 &23.31/0.738 \\
        \bottomrule
    \end{tabular*}
\end{table*}

\begin{table*}
    \caption{Quantitative comparisons on the PHSPD datasets.}
    \label{PHSPD comparisons}
    \centering
    \begin{tabular*}{1\textwidth}{@{\extracolsep{\fill}}@{\hspace{5pt}}c@{\hspace{5pt}}c@{\hspace{5pt}}c@{\hspace{5pt}}c@{\hspace{5pt}}c@{\hspace{5pt}}c@{\hspace{5pt}}c@{\hspace{5pt}}c@{\hspace{5pt}}c@{\hspace{5pt}}c@{\hspace{5pt}}c@{\hspace{5pt}}c@{\hspace{5pt}}c@{\hspace{5pt}}c@{\hspace{5pt}}@{}}
        \toprule
        \multirow{2}{*}{Methods} &\multirow{2}{*}{Params (M)} &\multirow{2}{*}{FLOPS (G)} &\multicolumn{4}{c}{PVFI-Mono}\\
         \cmidrule(r){4-7} 
         &&&Intensity		&Stokes		&AoLP		&DoLP \\
        \midrule
        CAIN \cite{choi2020channel}    &78.40 &\underline{19.94}&32.18/0.891&35.68/0.971&7.45/0.104&30.45/0.753\\
        FLAVR \cite{kalluri2023flavr}  &42.06 &99.98  &35.61/0.949 &39.31/0.984 &\textbf{9.21/0.271} &34.31/0.839\\
        VFIT-S \cite{shi2022video}     &\underline{7.54}     &29.83    &35.75/0.972 &40.26/0.989 &\underline{8.63/0.377} &\underline{38.46/0.935}\\
        VFIT-B \cite{shi2022video}     &29.08 &63.59 &\underline{36.33/0.972} &\underline{40.81/0.989} &8.46/0.358 &\textbf{38.75/0.933}\\
        Swin-VFI                       &\textbf{4.13}   &\textbf{18.33} &\textbf{37.24/0.976} &\textbf{41.66/0.991} &8.57/0.365 &38.17/0.932\\
        \bottomrule
    \end{tabular*}
\end{table*}

To further compare the performance of various networks on polarized data, this work train and test CAIN\cite{choi2020channel}, FLAVR\cite{kalluri2023flavr}, VFIT-S \cite{shi2022video},  VFIT-B \cite{shi2022video}  and our Swin-VFI on the PVFI-Mono dataset as well as the PHSPD dataset\cite{zou2022human}. Table \ref{PVFI-Mono comparisons} and Table \ref{PHSPD comparisons} report their performances of number of parameters (Params),  computational complexity (FLOPS), and peak signal-to-noise ratio (PSNR) and structural similarity index (SSIM) in terms of normalized Intensity, normalized Stokes, normalized AoLP and DoLP. As shown in the table, CAIN lags significantly behind the other methods in all the metrics. The performance of different models except CAIN on the two datasets shows the same pattern:  Swin-VFI  has a 45\% reduction in the number of parameters and a 39\% reduction in the number of FLOPS compared to the current lightest VFIT- S model. Thanks to the Transformer's ability to establish long-range correlations, it shows the best reconstruction accuracy in terms of intensity and Stokes, with an improvement of 0.32 dB and 0.91 dB in intensity and 0.46 dB and 1.02 dB in Stokes compared to FLAVR on both datasets, respectively. FLAVR performs best on the reconstruction of AoLP, with an improvement of 0.45 dB and 0.58 dB compared to the suboptimal method on the two datasets, respectively.VFIT- S and VFIT-B benefit from their spatio-temporal separation strategy, and perform best on the reconstruction of DoLP, with an improvement of 0.37 dB as well as 0.58 dB on the PSNR compared to Swin-VFI, respectively. 

\begin{figure*}[!h]
\centering
\includegraphics[width=1\textwidth]{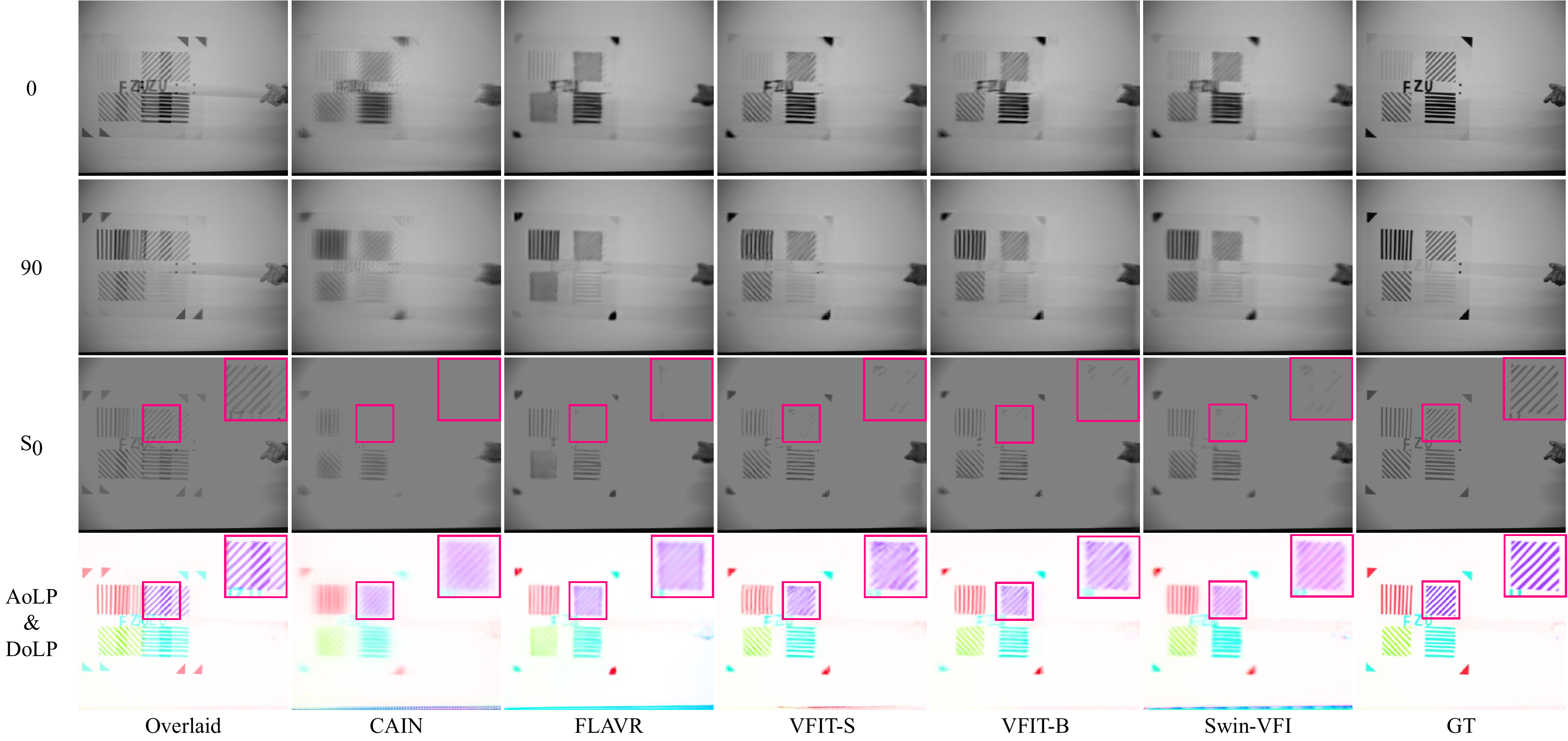}
\caption{Qualitative comparisons on the PVFI-Mono dataset with the state-of-the-art VFI methods in translation scenario. Note the hue differences within boxes for the comparison of DoLP and AoLP, which were mapped into saturation and hue separately (refer to the upper part of Fig. \ref{chanllenge} (b)).}
\label{translation}
\end{figure*}

\begin{figure*}[!h]
\centering
\includegraphics[width=1\textwidth]{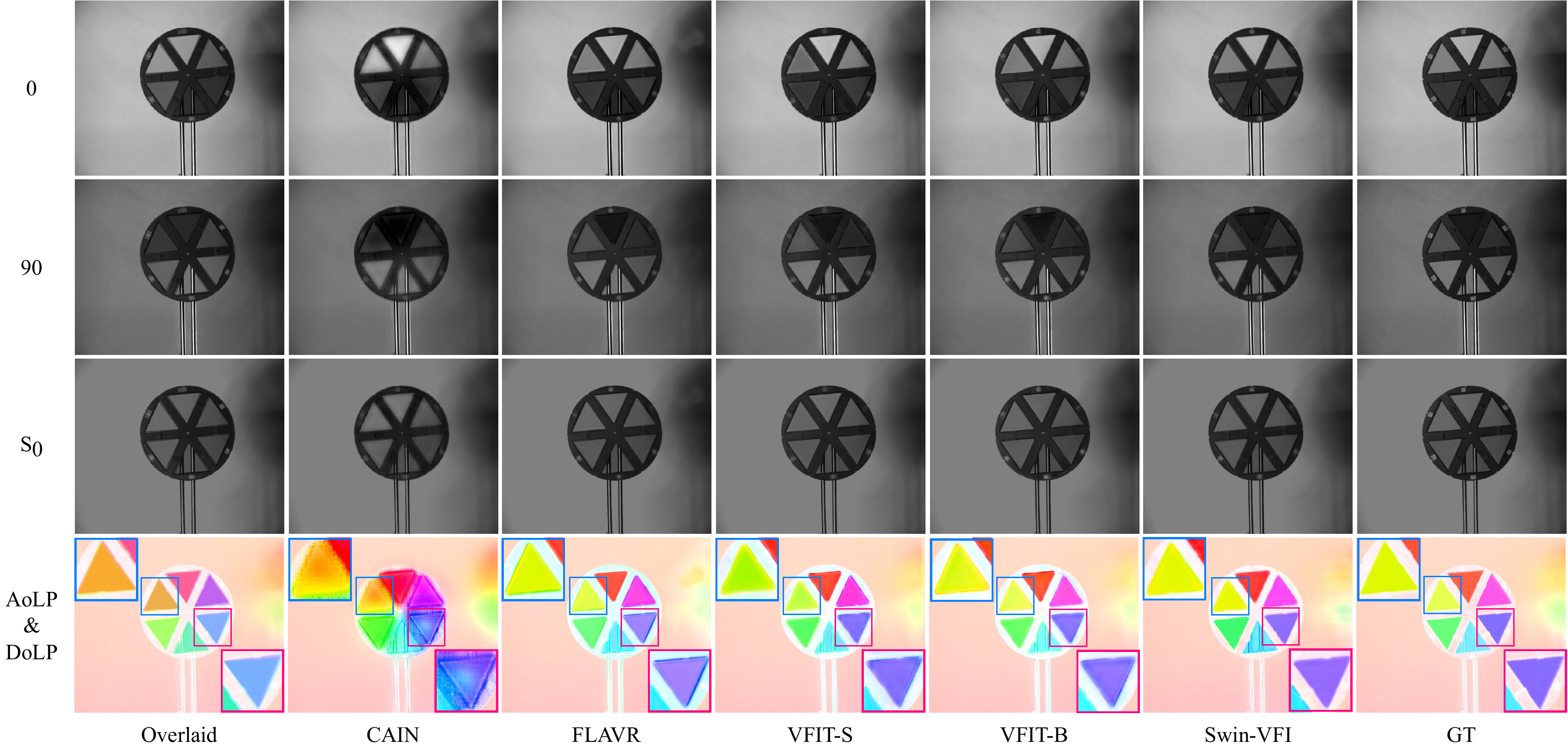}
\caption{Qualitative comparisons on the PVFI-Mono dataset with state-of-the-art VFI methods in rotation scenario. Note the hue differences within boxes for the comparison of DoLP and AoLP, which were mapped into saturation and hue separately (refer to the upper part of Fig. \ref{chanllenge} (b)).}
\label{rotation}
\end{figure*}

\begin{figure*}[!h]
\centering
\includegraphics[width=1\textwidth]{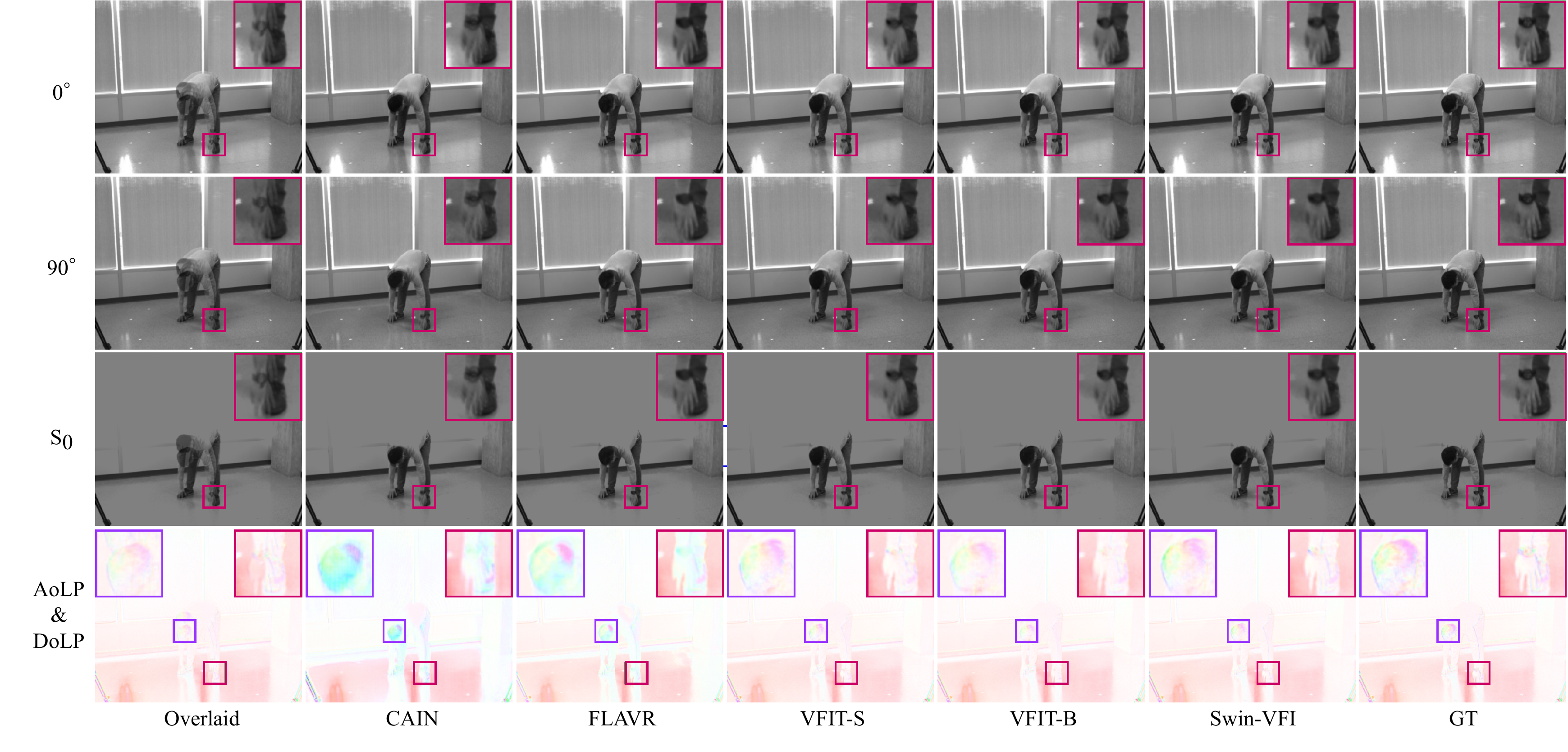}
\caption{Qualitative comparisons with state-of-the-art VFI methods using the PHSPD dataset. Note the hue differences within boxes for the comparison of DoLP and AoLP, which were mapped into saturation and hue separately (refer to the upper part of Fig. \ref{chanllenge}(b)) .}
\label{PHSPDcomparisons}
\end{figure*}

Fig. \ref{translation} and  Fig. \ref{rotation} illustrate the qualitative comparison of the different methods on the PVFI-Mono dataset in terms of 0, 90, normalized $\mathcal{S}_0$,  AoLP and DoLP. Consistent with the quantitative metrics, CAIN is inferior in both scenarios compared to the other methods. Due to the simple scene content of the PVFI-Mono dataset without much high-frequency information, the other methods other than CAIN did not show a significant difference in the intensity images at 0 and 90 degrees. In the translational motion scene, Swin-VFI exhibits clearer texture details in the magenta boxes of $\mathcal{S}_0$, AoLP and DoLP. In rotational motion scenes, Swin- VFI reconstructs more accurate AoLP and DoLP (note the hue differences within the magenta boxes for comparison).

\begin{figure*}[!h]
\centering
\includegraphics[width=1\textwidth]{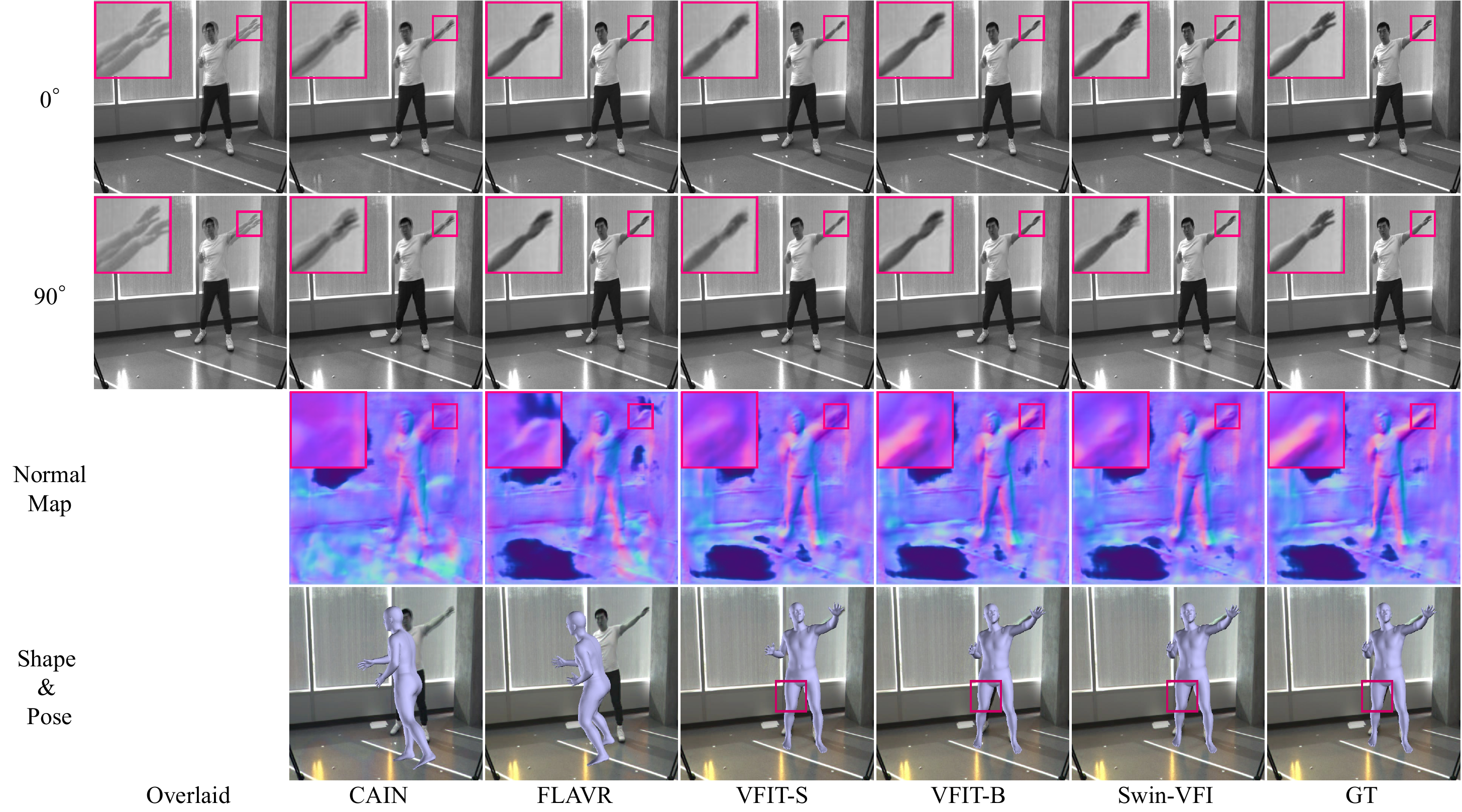}
\caption{Qualitative comparisons of practical performance of interpolated frames from different models in the tasks of computing surface normals and estimating human shape and pose.}
\label{PoseShape}
\end{figure*}

Fig. \ref{PHSPDcomparisons}  demonstrates the qualitative comparison of the different methods on the PHSPD dataset. As the results for 0, 90, $\mathcal{S}_0$ shown in the figure, Swin-VFI reproduces results with clearer structures. The AoLP and DoLP computed from the interpolated frames of CAIN and FLAVR show obvious errors in the head (purple box) and hand (magenta box) area. Compared to VFIT- S and VFIT-B, Swin-VFI restores more accurate AoLP and DoLP of the watch and clearer texture details of the fingers.

To assess the effectiveness of interpolated frames generated by different methods in practical tasks, such as estimating surface normal and human shape, this work utilized the method, data, and pre-trained models provided by \cite{zou2022human,zou2020detailed}. The original GT frames are replaced with the interpolated frames obtained from these methods to infer the normal map as well as the SMPL shape\cite{loper2023smpl} . As shown in Fig. \ref{PoseShape}, CAIN and FLAVR estimate reasonable surface normal but reconstruct inaccurate human pose, which could attribute to their weakness to provide accurate DoLP. The surface normal obtained from VFIT-B closely resembles that of GT, displaying a clear structure. However, upon observing the thigh region highlighted within the magenta box, subtle differences are noticeable between the pose derived from VFIT-S and VFIT-B in comparison to that obtained from the GT. Swin-VFI demonstrates a reconstruction of surface normals and human poses that closely align with those obtained from the GT.

\subsubsection{Comparisons on the conventional datasets}

This work further conduct a comparative analysis of Swin-VFI with methods mentioned above using conventional VFI datasets, including Vimeo-90K, Davis, SNU-FILM and Xiph. The performance of each model in terms of Params, PSNR, SSIM is reported in Table \ref{Vimeo comparisons} and Table \ref{SNU comparisons}. Compared with the current SOTA method VFIT-B on the Vimeo-90K dataset, Swin-VFI reduces the Params and FLOPS by 40\% and 39\% respectively, while achieving a notable performance improvement of 1.08 dB on Vimeo-90k testset and 0.96 dB on Davis testset. On SNU-FILM testset, Swin-VFI achieves a remarkable improvement of 1.59 dB in hard scenario, which indicates that Swin-VFI fully utilize the Transformer's ability to establish long-range correlations and prove their capability to handle challenging large-motion scenarios.

\begin{figure*}[!h]
\centering
\includegraphics[width=1\textwidth]{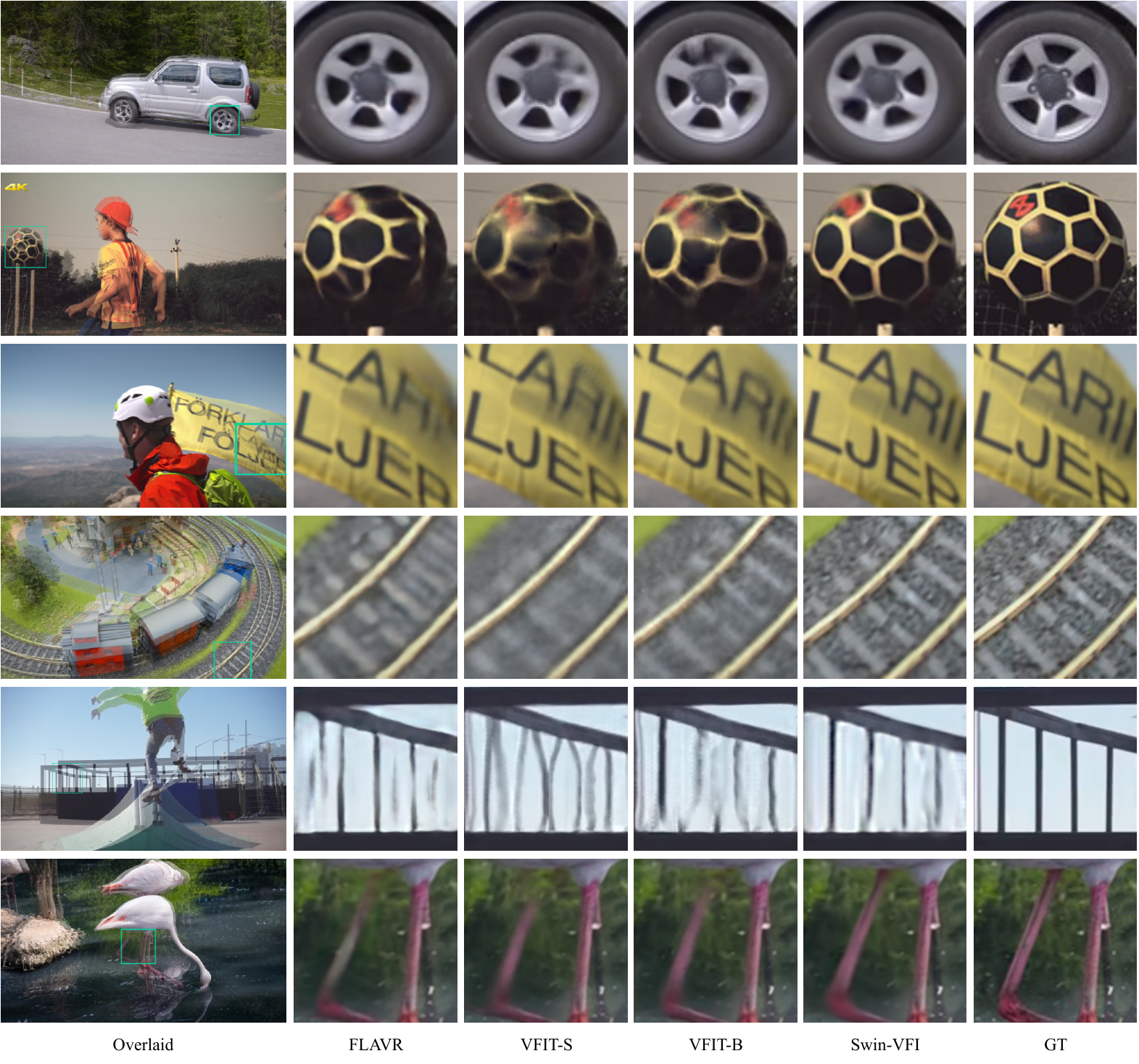}
\caption{Qualitative comparisons with state-of-the-art VFI methods using conventional VFI datasets. Swin-VFI outperforms others in providing precise motion estimation, clear texture details, handling non-rigid motion and large motion scenarios.}
\label{Qualitativecomparisons}
\end{figure*}

\begin{table*}[!h]
    \caption{Quantitative comparisons on the Vimeo-90K, DAVIS and Xiph datasets.}
    \label{Vimeo comparisons}
    \centering
    \begin{tabular*}{\textwidth}{@{\extracolsep{\fill}}@{\hspace{5pt}}c@{\hspace{5pt}}c@{\hspace{5pt}}c@{\hspace{5pt}}c@{\hspace{5pt}}c@{\hspace{5pt}}c@{\hspace{5pt}}c@{\hspace{5pt}}c@{\hspace{5pt}}c@{\hspace{5pt}}c@{\hspace{5pt}}c@{\hspace{5pt}}c@{\hspace{5pt}}c@{\hspace{5pt}}c@{\hspace{5pt}}@{}}
    % \begin{tabular*}{\textwidth}{ccccccccccc}
        \toprule
        \multirow{2}{*}{Methods}&\multirow{2}{*}{Params (M)}&\multirow{2}{*}{FLOPS (G)}  & \multirow{2}{*}{Vimeo-90K}  & \multirow{2}{*}{DAVIS} & \multicolumn{2}{c}{Xiph}\\
         \cmidrule(r){6-7}
         &&&&&2K &4K\\
        \midrule

        CAIN \cite{choi2020channel} & 42.78  &\textbf{11.21} & 34.83/0.970 & 27.21/0.873 &35.21/0.937 &32.56/0.901\\
        FLAVR \cite{kalluri2023flavr} &42.06&133.14 & 36.30/0.975 & 27.44/0.874 & 36.52/0.966 & 33.94/0.946\\
        VFIT-S \cite{shi2022video} &\textbf{7.54}		&\underline{40.09} & 36.48/0.976 & 27.92/0.885  & 37.07/0.968 & 34.36/0.948\\
        VFIT-B \cite{shi2022video} &29.08		&85.03	 & \underline{36.96/0.978} & \underline{28.09/0.888} & \underline{37.36/0.969}& \underline{34.57/0.949}\\
        % \textbf{SGuTA}  & \textbf{37.54/0.980} & \textbf{28.39/0.892} &\textbf{40.79/0.991}  &\textbf{37.41/0.985}   &\textbf{32.17/0.957}   &\textbf{26.15/0.880} & \textbf{37.52/0.969} & \textbf{34.83/0.950}\\
        \textbf{Swin-VFI}  &\underline{17.30}		&51.71		&\textbf{38.04/0.981}     	   &\textbf{28.86/0.899}&\textbf{38.14/0.971}&\textbf{35.36/0.952}\\
        \bottomrule
    \end{tabular*}
\end{table*}
\begin{table*}[!h]
    \caption{Quantitative comparisons on the SNU-FILM dataset.}
    \label{SNU comparisons}
    \centering
    \begin{tabular*}{\textwidth}{@{\extracolsep{\fill}}@{\hspace{5pt}}c@{\hspace{5pt}}c@{\hspace{5pt}}c@{\hspace{5pt}}c@{\hspace{5pt}}c@{\hspace{5pt}}c@{\hspace{5pt}}c@{\hspace{5pt}}c@{\hspace{5pt}}c@{\hspace{5pt}}c@{\hspace{5pt}}c@{\hspace{5pt}}c@{\hspace{5pt}}c@{\hspace{5pt}}c@{}}
    % \begin{tabular*}{\textwidth}{ccccccccccc}
        \toprule
        \multirow{2}{*}{Methods}&\multirow{2}{*}{Params (M)}&\multirow{2}{*}{FLOPS (G)}   &\multicolumn{4}{c}{SNU-FILM}\\
        \cmidrule(r){4-7} 
         &&&Easy		&Medium		&Hard		&Extreme\\
        \midrule

        CAIN \cite{choi2020channel} & 42.78  &\textbf{11.21}  &39.92/0.990		&35.61/0.978		&29.92/0.929		&24.81/0.851 \\
        FLAVR \cite{kalluri2023flavr} &42.06&133.14 &40.43/0.991		&36.36/0.981 		&30.86/0.942		&25.41/0.867 \\
        VFIT-S \cite{shi2022video} &\textbf{7.54}		&\underline{40.09}  &40.43/0.991		&36.52/0.983		&\underline{31.07/0.946}		&25.69/0.870 \\
        VFIT-B \cite{shi2022video} &29.08		&85.03	  &\underline{40.53/0.991} 		&\underline{36.53/0.982}		&31.03/0.945 	    &\underline{25.73/0.871} \\
        % \textbf{SGuTA}  & \textbf{37.54/0.980} & \textbf{28.39/0.892} &\textbf{40.79/0.991}  &\textbf{37.41/0.985}   &\textbf{32.17/0.957}   &\textbf{26.15/0.880} & \textbf{37.52/0.969} & \textbf{34.83/0.950}\\
        \textbf{Swin-VFI}  &\underline{17.30}		&51.71		&\textbf{40.90/0.992}	&\textbf{37.78/0.986}	&\textbf{32.62/0.960} &\textbf{26.37/0.884} \\
        \bottomrule
    \end{tabular*}
\end{table*}
This study provide qualitative results comparing Swin-VFI model to FLAVR \cite{kalluri2023flavr} and VFIT \cite{shi2022video}. As shown in Fig. \ref{Qualitativecomparisons}. The first two rows fully demonstrate the ability of Swin-VFI to provide accurate motion estimation (Please carefully compare the rotation of the wheels and balls with the ground truth; other methods fail to restore the accurate rotation angles). The third row shows the performance of various models in non-rigid motion scenarios, where only Swin-VFI clearly restores all the letters. In the fourth row, Swin-VFI reconstruct clearer texture details. The last two rows demonstrate the strong ability of our models to handle large motion scenarios.
\subsection{Ablation Study}

\begin{table}[h]
\centering
    \caption{Ablation study of $\mathcal{L}$}
    \begin{tabular}{@{\extracolsep{\fill}}@{\hspace{5pt}}c@{\hspace{5pt}}c@{\hspace{5pt}}c@{\hspace{5pt}}c@{\hspace{5pt}}c@{}}
    % \begin{tabular}{ccccc}
    \toprule

 $\mathcal{L}$  &Intensity     &Stokes	&AoLP 	&DoLP\\
    \midrule
    $1 \mathcal{L}_I$                               &\textbf{28.23/0.883}&31.60/0.951&10.27/0.298&23.14/0.738\\
    $1 \mathcal{L}_S$                               &20.94/0.380&30.57/0.940&10.23/0.267&22.21/0.682 \\
    $0.1 \mathcal{L}_I + 1\mathcal{L}_S$          &28.17/0.880 &\underline{31.91/0.953}&10.28/0.297&\underline{23.31/0.738} \\
    $0.9  \mathcal{L}_I + 0.1 \mathcal{L}_A $         &27.85/0.872&31.40/0.952&\textbf{11.44/0.308}&23.00/0.723\\
    $0.7  \mathcal{L}_I + 0.3 \mathcal{L}_D$          &\underline{28.23/0.882}&\textbf{31.94/0.953}&\underline{10.30/0.298}&\textbf{23.37/0.747}\\
    \bottomrule

    \end{tabular}
    
    \label{Loss}
\end{table}
\subsubsection{Loss function}
Table \ref{Loss} presents the performance of different loss functions obtained with setting of 
$\lambda$ in Eq. \ref{L} on the PVFI-Mono dataset.
As shown in the table, in the scenario where the loss solely consists of the Stokes term, all metrics exhibit varying degrees of degradation, indicating the necessity of the Intensity term. Upon incorporating either the AoLP or DoLP term on the Intensity term, a noticeable enhancement in the corresponding reconstruction accuracy is observed. However, the introduction of the AoLP constraint leads to a decline in the reconstruction accuracy of Intensity, Stokes, and DoLP. This degradation is not evident when adding the DoLP term, showcasing optimal performance in reconstructing Stokes and DoLP, while exhibiting suboptimal performance in Intensity and AoLP. Despite its superior overall performance, challenges such as gradient explosion and training instability are observed when employing Intensity and DoLP as the loss function in training on other methods. This issue arises due to the calculated pixel values of $S_0$ being close to zero at certain pixel positions during the interpolation of frames, as referenced in Eq. \ref{adolp}. At these pixel positions, the DoLP can be significantly greater than 1, resulting in anomalies in the DoLP term's loss. Given that the results obtained by adding the Stokes term only slightly lag behind the scenario with the addition of the DoLP term in terms of reconstruction accuracy, this study ultimately set $\lambda_1=0.1,\lambda_2=1,\lambda_3=\lambda_4=0$ in Eq. \ref{L}.
\label{loss}

% \begin{table}[ht]
% \centering
%     \caption{Ablation study of $\mathcal{L}$}
%     \begin{tabular}{@{\extracolsep{\fill}}@{\hspace{5pt}}c@{\hspace{5pt}}c@{\hspace{5pt}}c@{\hspace{5pt}}c@{\hspace{5pt}}c@{}}
%     % \begin{tabular}{ccccc}
%     \toprule

%  $\mathcal{L}$  &Intensity     &Stokes	&AoLP 	&DoLP\\
%     \midrule
%     $1 \mathcal{L}_I$                               &28.60/0.887      &32.37/0.957      &10.18/0.296      &23.56/0.727\\
%     $0.1  \mathcal{L}_I + 1  \mathcal{L}_S$          &\textbf{29.06/0.896}      &\textbf{32.88/0.959}        &10.60/0.313     &24.21/0.746\\
%     $1  \mathcal{L}_I + 0.5 \mathcal{L}_A $          &26.21/0.802      &29.49/0.931       &\textbf{11.60/0.263}     &20.33/0.635\\
%     $0.4  \mathcal{L}_I + 0.6 \mathcal{L}_D$          &29.01/0.896       &32.80/0.958       &10.51/0.312     &\textbf{24.25/0.752}\\

%     \bottomrule

%     \end{tabular}
%     % \subcaption{Table 2}
    
%     \label{Loss}
% \end{table}
\subsubsection{Self-Attention Mechanism}
\label{Self-Attention Mechanism}

In this section, various self-attention mechanisms applicable to VFI task are compared. To ensure fairness, all methods are configured with $N_s=2$. We firstly replace all Swin-T blocks with two layers of 3D ResBlocks \cite{he2016deep} to enable a comparative assessment of CNN-based methods with other Transformer-based methods. Subsequently, a thorough evaluation of the performance of different MSAs is conducted. Baseline removed all MSA modules of Swin-T block. Global MSA, Channel MSA \cite{cai2022mst++}, STS MSA\cite{dosovitskiy2020image} and Sep-STS MSA \cite{shi2022video} replace the MSA layer in Swin-T block with ViT, SAB, STS and Sep-STS layers respectively. Swin-T is the method employed in this paper. 

\begin{table}[h]
  % \begin{adjustbox}{minipage=0.48\linewidth, valign=t}
    \centering
    \caption{Ablation study of different MSA on VFI task}
    % \begin{tabular}{@{}c@{\hspace{2pt}}c@{\hspace{6pt}}c@{\hspace{4pt}}c@{}}
    \begin{tabular}{cccc}

    \toprule
    Methods        &Params (M)	&FLOPS (G)	&Vimeo-90K\\
    % \multirow{2}{*}{Methods}        &Params	&FLOPS	&\multicolumn{2}{c}{Vimeo-90K}\\
    % \cline{4-5}
                                %  &(M)                 &(G)               &PSNR   &SSIM    \\
    \midrule
    3D ResBlock\cite{he2016deep}     &17.50        &57.94        &35.95/0.974   \\
    Baseline        &\textbf{7.84}&\textbf{5.12}&36.92/0.977  \\
    Global\cite{vaswani2017attention}       &8.76         &587.93       &-\\
    Channel MSA\cite{cai2022mst++}         &8.76       &22.11        &36.95/0.977\\
    STS MSA\cite{shi2022video}         &11.53        &47.97        &37.32/0.979\\
    Sep-STS MSA\cite{shi2022video}    &10.61        &29.76        &37.13/0.978\\ 
    % SGuTA           &18.40        &49.04        &37.28/0.979\\
    Swin-VFI           &11.53      &34.48       &\textbf{37.48/0.980}\\
    \bottomrule
  \end{tabular}
    % \subcaption{Table 1}
    \label{Comparison of different MSA}
    \end{table}

As shown in Table \ref{Comparison of different MSA}, on the one hand, compared to the 3D ResBlock method based on CNN, Transformers benefit from their ability to establish long-range dependencies, achieving improved performance with lower Params and FLOPS. On the other hand, Compared to the Baseline, Global predictably encountered an "Out of Memory" issue during the training process. Channel MSA only provides a modest improvement in PSNR by 0.06 dB, indicating that the self-attention for channels has limited benefit for VFI task. STS MSA and Sep-STS MSA show PSNR improvements of 0.40 dB and 0.21 dB, respectively, with Sep-STS MSA acting similarly to depth-wise separable convolution \cite{howard2017mobilenets}, resulting in a lighter and more efficient STS-MSA. Swin-VFI leverages the shifted-cube mechanism to fully exploit the power of local attention, achieving the best performance among all MSAs.

\subsubsection{Stage}
\begin{table}[h]
\centering
    \caption{Ablation study of stage number}
    \begin{tabular}{ccccc}
    \toprule
    % \multirow{2}{*}{Methods}  &\multirow{2}{*}{$N_s$}      &Params	&FLOPS	&\multicolumn{2}{c}{Vimeo-90K}\\
    Dataset  &$N_s$     &Params(M)	&FLOPS(G) 	&PSNR/SSIM\\
    \midrule
    \multirow{3}{*}{PVFI-Mono}            &1       &\textbf{4.13}        &\textbf{18.33}      &28.17/0.880\\
                &2       &8.27      &36.65      &\textbf{28.27/0.881}\\ 
                &3       &12.40  &54.97      &28.26/0.880 \\

    \midrule
    \multirow{3}{*}{Vimeo-90K}            &1       &\textbf{5.77}        &\textbf{17.24}      &36.72/0.976\\
            &2        &11.53       &34.48      &37.48/0.980\\
            &3        &17.30       &51.71      &\textbf{38.04/0.981}\\ 
    \bottomrule

    \end{tabular}
    % \subcaption{Table 2}
    
    \label{stage number}
\end{table}
In this section,  the impact of stage number $N_s$ is explored. Due to concerns regarding Params and FLOPS, this study only consider cases when $N_s \leq 3$. The results presented in Table \ref{stage number} reveal that, on one hand, for the PVFI-Mono dataset, although the model exhibits the best performance with $N_s=2$, the improvement is limited. For the sake of model lightness and efficiency, $N_s$ is set to 1 for the polarization data. On the other hand, for the Vimeo-90K dataset, Swin-VFI performs the best when $N_s=3$. Additionally, it is worth noting that when $N_s=1$, compared to VFIT-S, Swin-VFI achieves a PSNR improvement of 0.24 dB while reducing the Params and FLOPS by 23\% and 58\%, respectively. When $N_s=2$, compared to the current SOTA method VFIT-B, Swin-VFI achieves a PSNR improvement of 0.52 dB with 43\% of its Params and 40\% of the FLOPS.

\section{Conclusion}
This study represents the inaugural endeavor to tackle the polarized video frame interpolation task through a data-driven approach. The necessity and challenges of this task were explored comprehensively. To facilitate a more focused analysis and mitigate potential confounding factors, a dataset named PVFI-Mono, characterized by strong polarization information within a simple motion scenario, was curated. Reconstruction accuracy was enhanced by integrating local attention into a multi-stage multi-scale framework, resulting in Swin-VFI. This method leverages the transformer's capability to establish long-range correlations while maintaining relatively lower computational complexity. Additionally, a polarization loss term was introduced to augment our approach, improving the reconstruction accuracy of polarization information. Both qualitative and quantitative results attest to the superiority of our method over recent state-of-the-art approaches in both polarized VFI and conventional VFI tasks. Superior reconstruction accuracy in terms of intensity and Stokes parameters for polarized video demonstrates the successful applicability of our method to SfP and Human Shape Reconstruction tasks. Future endeavors will extend to exploring the color polarized video frame interpolation task, building upon the foundations laid in polarized VFI and conventional VFI studies.

%%%%%%%%%%%%%%%%%%%%%%% References %%%%%%%%%%%%%%%%%%%%%%%%%
\begin{backmatter}
%\bmsection{Funding}
%The work was supported by the Natural Science Foundation of Fujian Province of China (No. 2023J01130137).
 \bmsection{Disclosures}
 The authors declare no conflicts of interest.
 \bmsection{Data availability} Data underlying the results presented in this paper are not publicly available at this time but may be obtained from the authors upon reasonable request.	
\end{backmatter}

%%%%%%%%%% If using BibTeX:
\bibliography{sample}

%%%%%%%%%% If preparing manually:
% \begin{thebibliography}{1}
% \newcommand{\enquote}[1]{``#1''}

% \bibitem{Zhang:14}
% Y.~Zhang, S.~Qiao, L.~Sun, Q.~W. Shi, W.~Huang, L.~Li, and Z.~Yang,
%   \enquote{Photoinduced active terahertz metamaterials with nanostructured
%   vanadium dioxide film deposited by sol-gel method,}
%   {\protect\JournalTitle{Optics Express}} \textbf{22}, 11070--11078 (2014).

% \bibitem{Optica}
% {Optica}, \enquote{{Optica Publishing Group},}
%   \url{http://www.opg.optica.org}.

% \bibitem{FORSTER2007}
% P.~Forster, V.~Ramaswamy, P.~Artaxo, T.~Bernsten, R.~Betts, D.~Fahey,
%   J.~Haywood, J.~Lean, D.~Lowe, G.~Myhre, J.~Nganga, R.~Prinn, G.~Raga,
%   M.~Schulz, and R.~V. Dorland, \enquote{Changes in atmospheric consituents and
%   in radiative forcing,} in \enquote{Climate Change 2007: The Physical Science
%   Basis. Contribution of Working Group 1 to the Fourth assesment report of
%   Intergovernmental Panel on Climate Change,}  S.~Solomon, D.~Qin, M.~Manning,
%   Z.~Chen, M.~Marquis, K.~B. Averyt, M.~Tignor, and H.~L. Miler, eds.
%   (Cambridge University Press, 2007).

% \end{thebibliography}

\end{document}